%% file: main.tex
\documentclass[peerreview]{IEEEtran}
\usepackage[utf8]{inputenc}
\usepackage{subcaption}
\usepackage{amsmath}

\usepackage{cite} 
\usepackage{url} 
\usepackage[utf8]{inputenc} 
\usepackage{booktabs} 
\usepackage{graphicx}
\usepackage{authblk}

\usepackage{xcolor}
\usepackage[linesnumbered,ruled,vlined]{algorithm2e}

\usepackage{multirow}
\usepackage{amsfonts} 

\newtheorem{definition}{Definition}[section]
\newtheorem{example}{Example}
\DeclareMathOperator*{\argmax}{arg\,max}
\DeclareMathOperator*{\argmin}{arg\,min}

\usepackage{wrapfig}
\usepackage{lipsum} 

\begin{document}
\title{A Report on Semantic-Guided RL for \\  Interpretable Feature Engineering}

\author[1]{Mohamed Bouadi}
\author[2]{Arta Alavi}
\author[1]{Salima Benbernou}
\author[1]{Mourad Ouziri}
\affil[1]{Université Paris Cité, LIPADE}
\affil[2]{SAP Labs Paris}

\maketitle
\tableofcontents
\listoffigures
\listoftables

\IEEEpeerreviewmaketitle
\begin{abstract}
The quality of Machine Learning (ML) models strongly depends on the input data, as such generating high-quality features is often required to improve the predictive accuracy. This process is referred to as Feature Engineering (FE). However, since manual FE is time-consuming and requires case-by-case domain knowledge, AutoFE is crucial. A major challenge that remains is to generate interpretable features. To tackle this problem, we introduce \textit{SMART}, a hybrid approach that uses semantic technologies to guide the generation of interpretable features through a two-step process: Exploitation and Exploration. The former uses Description Logics (DL) to reason on the semantics embedded in Knowledge Graphs (KG) to infer domain-specific features, while the latter exploits the KG to conduct a guided exploration of the search space through Deep Reinforcement Learning (DRL). Our experiments on public datasets demonstrate that \textit{SMART} significantly improves prediction accuracy while ensuring a high level of interpretability. 

\end{abstract}

\section{Introduction}
\begin{figure}[ht!]
    \centering
    \includegraphics[width=1\linewidth, height=4cm]{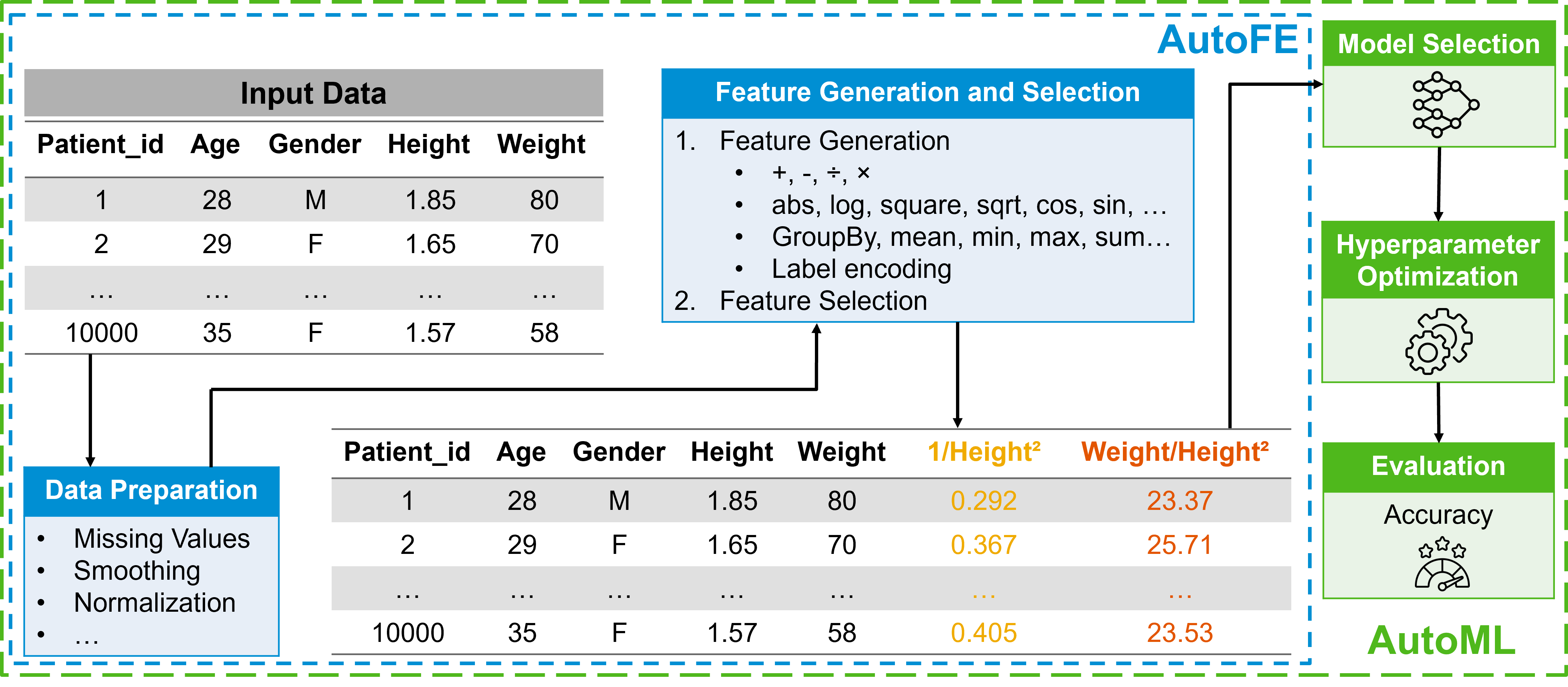}
    \caption{The different workflows of AutoFE and AutoML.}
    \label{fig:example}
\end{figure}

Nowadays, ML has demonstrated impressive results with structured data \cite{li2022camel,DBLP:journals/pacmmod/GuFCL0W23}. Its success is often attributed to the experience of data scientists who leverage domain knowledge to extract useful patterns from data. This crucial, yet tedious task, is commonly known as Feature Engineering (FE). In practice, data scientists rely on their domain knowledge to identify useful features in a trial-and-error manner, which requires case-by-case domain knowledge. However, given the increasing computational capabilities, AutoFE emerges as an important topic in AutoML, as shown in Fig. \ref{fig:example}, as it may reduce data scientists workload for quicker decision-making \cite{DBLP:journals/pvldb/00080LWZCDHWWZR21}. Nevertheless, AutoML tools are still focusing on model selection and hyper-parameter optimization and often neglecting FE.

\begin{example} (Feature Generation).
Consider the problem of predicting heart disease of patients based on their attributes such as height, weight, age, amongst others. While these raw features provide valuable information, a more useful feature for this problem would be the \textit{Body Mass Index}, $BMI = \frac{weight}{height^2}$, which can be derived using two functions – \textit{square} and \textit{division}, as shown in Fig. \ref{fig:example}. 
\end{example}

Recently, several approaches of AutoFE have been proposed \cite{khurana2018feature,chen2019neural,waring2020automated}. However, these approaches have several drawbacks. For instance, they suffer from performance bottlenecks due to the large number of candidates. Additionally, they lack the ability to transfer learning, i.e., they fail to leverage prior knowledge to minimize the exploration phase when dealing with unseen data, a capability inherent to human experts. Furthermore, these tools fall short in using external knowledge that is usually embedded in Ontologies and Knowledge Graphs \cite{DBLP:conf/semweb/DashBMG23,DBLP:conf/semweb/AsprinoC23}.

Interpretability has been an important goal since the early days of AI. Recent studies have shown that the interpretability of ML models (IML) strongly depends on the interpretability of the input features \cite{zytek2022need,roscher2020explainable}. This highlights the importance of feature interpretability, not only for improved performance but also to enhance trust of domain experts in ML models \cite{DBLP:conf/sigmod/PradhanZGS22,gunning2019xai}. Nevertheless, existing AutoFE methods often struggle to discover easily understandable and interpretable features. This is mainly due to the lack of research formalizing what makes a feature interpretable, which usually refers to features that are human-generated \cite{yadav2017explanation}, and almost no work quantifying feature interpretability, despite the agreement on its importance in the literature \cite{sokol2018glass,choi2020ten}.

Authors in \cite{preece2018stakeholders} introduced different stakeholders in explainable AI (XAI), amongst them \textit{domain experts}, who use ML models to make important decisions. These users usually require feature interpretability to trust the model's output \cite{zytek2022need}. Therefore, there is a real need for an approach that automates the generation of interpretable features for domain experts, enabling them to gain insights into predictive and data mining problems.

In recent years, KG along with reasoning mechanisms have gained widespread adoption in a large number of domains \cite{DBLP:conf/semweb/IglesiasMolinaAIPC23,DBLP:conf/semweb/WuWLZWDHLR23}. KG and logical rules can provide context and semantic information about features and their valid combinations. This can be done through leveraging domain knowledge and linking data items with the semantic concepts and instances of the KG.

In this work, we introduce a semantic-guided RL solution to simultaneously maximize model performance and generate interpretable features for domain experts. We use DL to reason over a KG to generate interpretable features through a two-step process: \textit{Exploitation} and \textit{Exploration}. The former leverages the KG to infer relevant features using a reasoner, while the latter, employs DRL, more specifically, Deep Q-Network algorithm (DQN), to iteratively generate features through interacting with the environment. We introduce a novel metric to evaluate the interpretability of newly generated features based on their relationships with the entities of the KG. To the best of our knowledge, our work is the first to address the trade-off between model accuracy and feature interpretability.  

\subsubsection*{\textbf{Contributions.}} We summarize our contributions as follows: 

\begin{enumerate}

    \item We define feature interpretability in ML, taking into account both features semantics and structure.
    
    \item We formalize FE problem as a multi-optimization problem that maximizes both feature interpretability and the ML model's accuracy.

    \item We propose a semantic-based solution that takes advantage of the semantics embedded in the domain knowledge to automatically engineer interpretable features using a DRL agent.
    
    \item We conduct a set of experiments on a large number of public datasets to validate our solution on both metrics: performance and interpretability.
\end{enumerate}
The report is organized as follows. In Section 2, we review the related work. Section 3 provides the data model and the problem definition. In Section 4, we present our solution for interpretable FE and, in section 5, we subject our approach to a set of experiments. Finally, Section 6 concludes the report.

\section{Problem Definition}

In this work, we address the problem of engineering interpretable features using logical reasoning mechanisms based on the knowledge embedded in a dataset. Our approach involves identifying effective transformations based on a given set of input features. 

\subsection{Defining Interpretability}
Building a model to generate interpretable features requires a clear definition of interpretability. However, there is no consensus regarding the definition of interpretability in ML and how to evaluate it. Since a formal definition could be elusive, we have sought insights from the field of psychology.

\begin{definition}(Interpretability). In general, to interpret means "to explain the meaning" or "to present in understandable terms". In psychology \cite{lombrozo2006structure}, interperetability refers to the ability to understand and make sense of an observation. In the context of ML, the authors in \cite{doshi2017towards} defines interpretability as "the ability to explain or to present in understandable terms to a human".
\end{definition}

ML needs to learn from good data. Even simple, interpretable models, like regression, become difficult to understand with non-interpretable features. However, different users may have different requirements when it comes to feature interpretability. Presently, literature is focusing on models interpretability, without necessarily considering whether the input features are interpretable for domain experts. To this end, in our research, we are focusing on \textit{feature interpretability} for \textit{domain experts}, by capturing feature structure and semantics, along with their connections to the domain knowledge.

\begin{definition}(Feature Interpretability). We define feature interpretability as the intellectual effort exerted by domain experts to understand and make sense of the generated features aligning them with their domain knowledge. In essence, interpretability involves the effort of mapping the features to relevant concepts and semantics within the domain of interest to gain deeper insights into the training data of ML models.
\label{def:interp}
\end{definition}

Consequently, interpretable features should be humanly-readable and refer to real-world entities  that domain experts can understand and reason about. 

\subsection{Feature Engineering Pipeline}
Consider a predictive problem on a tabular dataset $D=(X,Y)$ consisting of: 

\begin{enumerate}
    \item A set of features $X=\{x_1,..,x_p\} \in \mathbb{R}^{n \times p}$, where $n$ is the number of instances and $p$ is the number of features;
    
    \item An applicable ML model, $L$, (e.g. Random Forest); 
    
    \item A corresponding cross-validation performance measure $\mathcal{P}$ (e.g. F1-score); 
    
    \item An interpretability function, $\mathcal{I}_{KG} : Dom^F \rightarrow [0,1]$, where $Dom^F$ is the set of all possible features, that returns an interpretability score of a feature $x \in X$ based on Definition \ref{def:interp}. 
\end{enumerate}
We define a FE pipeline $\mathcal{T} = \{t_1, ..., t_m\}$ as an ordered sequence of $m$ transformations applied to $X$. The set of generated features from $X$ using $\mathcal{T}$ is denoted as $\hat{X}_{\mathcal{T}}$.

The goal of AutoFE is to find the optimal FE pipeline, $\mathcal{T}$, that generates $\hat{X}_{\mathcal{T}}$ which maximizes the performance $\mathcal{P}(L(\hat{X}_{\mathcal{T}},Y))$ for a given algorithm $L$ and a metric $\mathcal{P}$, as shown in Equation \ref{eq:FE1}:

\begin{equation}
    \label{eq:FE1}
    \begin{split}
        \mathit{\mathcal{T} = \argmax_{\mathcal{T}} \mathcal{F}(\mathcal{P}(L(\hat{X}_{\mathcal{T}},Y)),\mathcal{I}_{KG}(\hat{X}_{\mathcal{T}}))} \\
        \noindent s.t.\\
        \mathit{\mathcal{I}_{KG}(\hat{X}_{\mathcal{T}}) =\sum_{\hat{x_i} \in \hat{X}_{\mathcal{T}}}\mathcal{I}_{KG}(\hat{x_i})} \\
    \end{split}
\end{equation}
where $\mathcal{F}$ represents a bi-objective function that assesses both model performance, $\mathcal{P}(L(\hat{X}_{\mathcal{T}},Y))$, and feature interpretability $\mathcal{I}_{KG}(\hat{X}_{\mathcal{T}})$. 

To solve this bi-objective optimization problem, we used Scalarization technique \cite{trummer2014approximation,benouaret2021multi}, which combines the two objectives into a single one with a weighted linear combination. 

\begin{equation}
    \label{eq:FE2}
    \begin{split}
        \mathit{\mathcal{T} = \argmax_{\mathcal{T}} (\lambda \mathcal{P}(L(\hat{X}_{\mathcal{T}},Y))+(1-\lambda)\mathcal{I}_{KG}(\hat{X}_{\mathcal{T}}))} \\
        \noindent s.t.\\
        \mathit{\mathcal{I}_{KG}(\hat{X}_{\mathcal{T}}) =\sum_{\hat{x_i} \in \hat{X}_{\mathcal{T}}}\mathcal{I}_{KG}(\hat{x_i})} \\
        \mathit{\lambda, \mathcal{I}_{KG}(\hat{x_i}) \in [0,1] \forall{\hat{x_i} \in \hat{X}_{\mathcal{T}}}}\\
    \end{split}
\end{equation}

\section{Proposed Solutions}
To address the challenges of engineering interpretable features, we propose \textit{SMART}. We first present an overview (Fig. \ref{fig:Archi}), then we elaborate on the components.

\subsection{An Overview of \textit{SMART}}
\begin{figure}[ht!]
    \centering
    \includegraphics[width=1\linewidth,height=5cm]{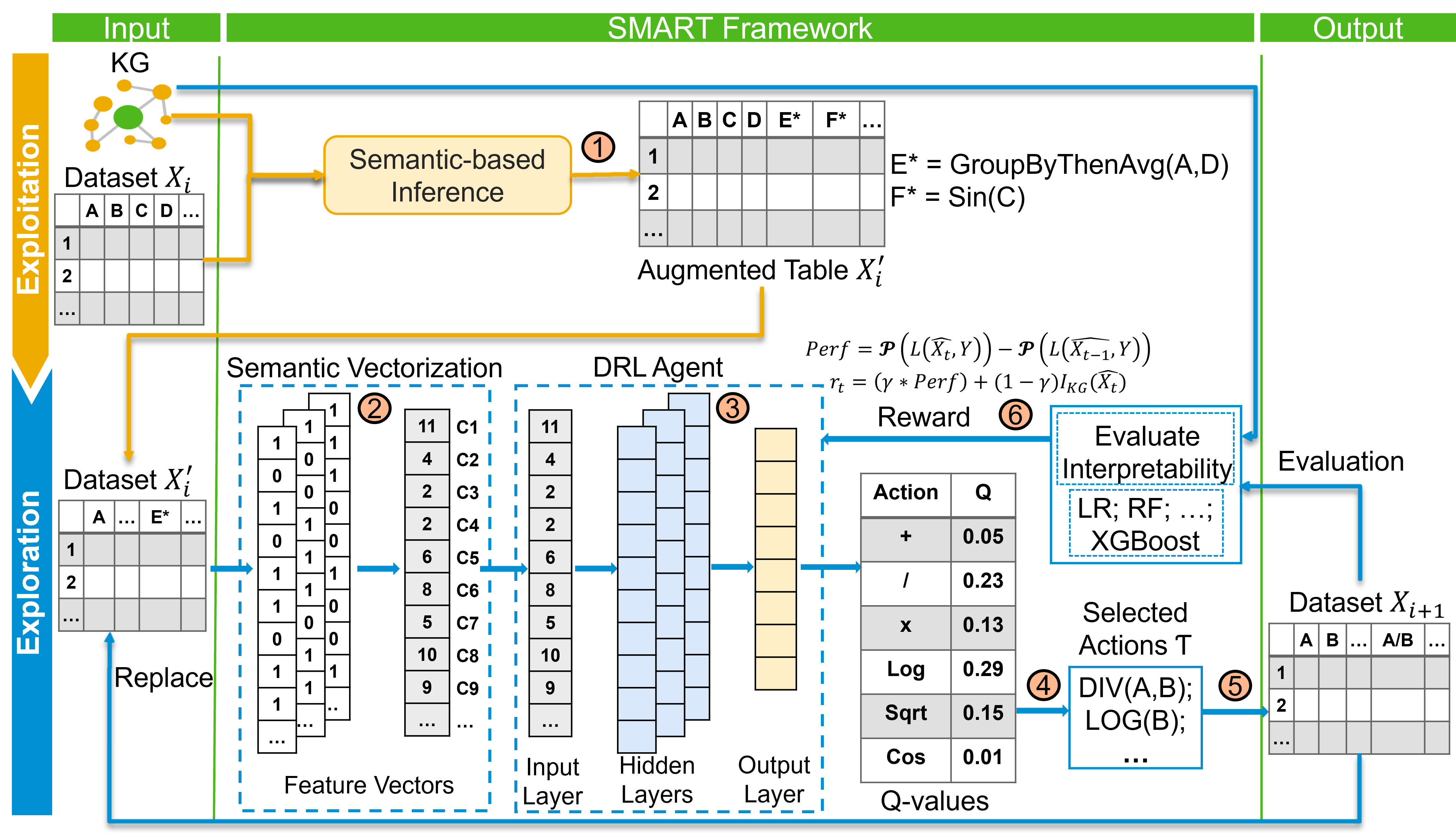}
    \caption{An overview of \textit{SMART} architecture.}
    \label{fig:Archi}
\end{figure}

\textit{SMART} consists of two main steps:\textit{Exploitation} and \textit{Exploration} (Fig. \ref{fig:Archi}). During the \textit{Exploitation}, we first employ a semantic mapping to identify target entities from the KG that are related to the input data. We then perform various reasoning tasks to infer relationships within the KG to generate new features, resulting in a richer feature set that will be the input of the second step (Fig. \ref{fig:Archi}.1). During \textit{Exploration}, we perform a semantic vectorization to represent the input data with a feature vector based on the KG semantics (Fig. \ref{fig:Archi}.2). Then, we used a DQN to estimate the transformations probabilities based on historical data (Fig. \ref{fig:Archi}.3). The transformations with higher probability are used to generate new features (Fig. \ref{fig:Archi}.4) and the agent is rewarded accordingly. The reward  (Fig. \ref{fig:Archi}.5) is a combination of the effectiveness of the new features, calculated by training a model on the new dataset, and the interpretability score of the generated features, determined using a new metric based on Definition \ref{def:interp}. This process continues iteratively until convergence of the model.

\input{Exploitation}

\input{Exploration}

\input{Exp.tex}

\section{Conclusion}
Our work aims at bridging the gap between statistical AI, i.e., ML, and symbolic AI that can be applied in many fields such as data management for decision making or data mining. Along these lines, we presented a novel technique to efficiently engineer interpretable features from structured knowledge. The cornerstone of our approach are - a human-like logical reasoner to exploit a KG, thereby enriching the feature set with interpretable features and a DRL-based exploration of the feature space to enhance the model performance while maintaining a high level of interpretability. Extensive experiments on large scale datasets are conducted, and detailed analysis is provided, which shows that \textit{SMART} can provide competitive performance and guarantee the interpretability of the generated features.

\bibliographystyle{ACM-Reference-Format}
\bibliography{main}

\end{document}

%% file: Exploitation.tex
\subsection{Exploitation}
In this step, \textit{SMART} leverages a KG to enrich the input dataset with new semantic-based features using Description Logics.

\begin{figure}
    \centering
    \captionsetup[subfigure]{labelformat=empty}
    
    \begin{subfigure}{0.49\linewidth}
        \centering
        \includegraphics[width=\linewidth, height=3cm]{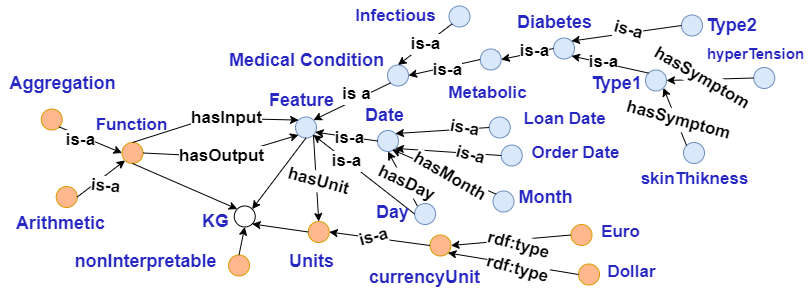}
        \caption{\textit{SMART}\_KG}
        \label{fig:kg}
    \end{subfigure}
    \hfill
    \begin{subfigure}{0.49\linewidth}
        \centering
        \includegraphics[width=\linewidth, height=3cm]{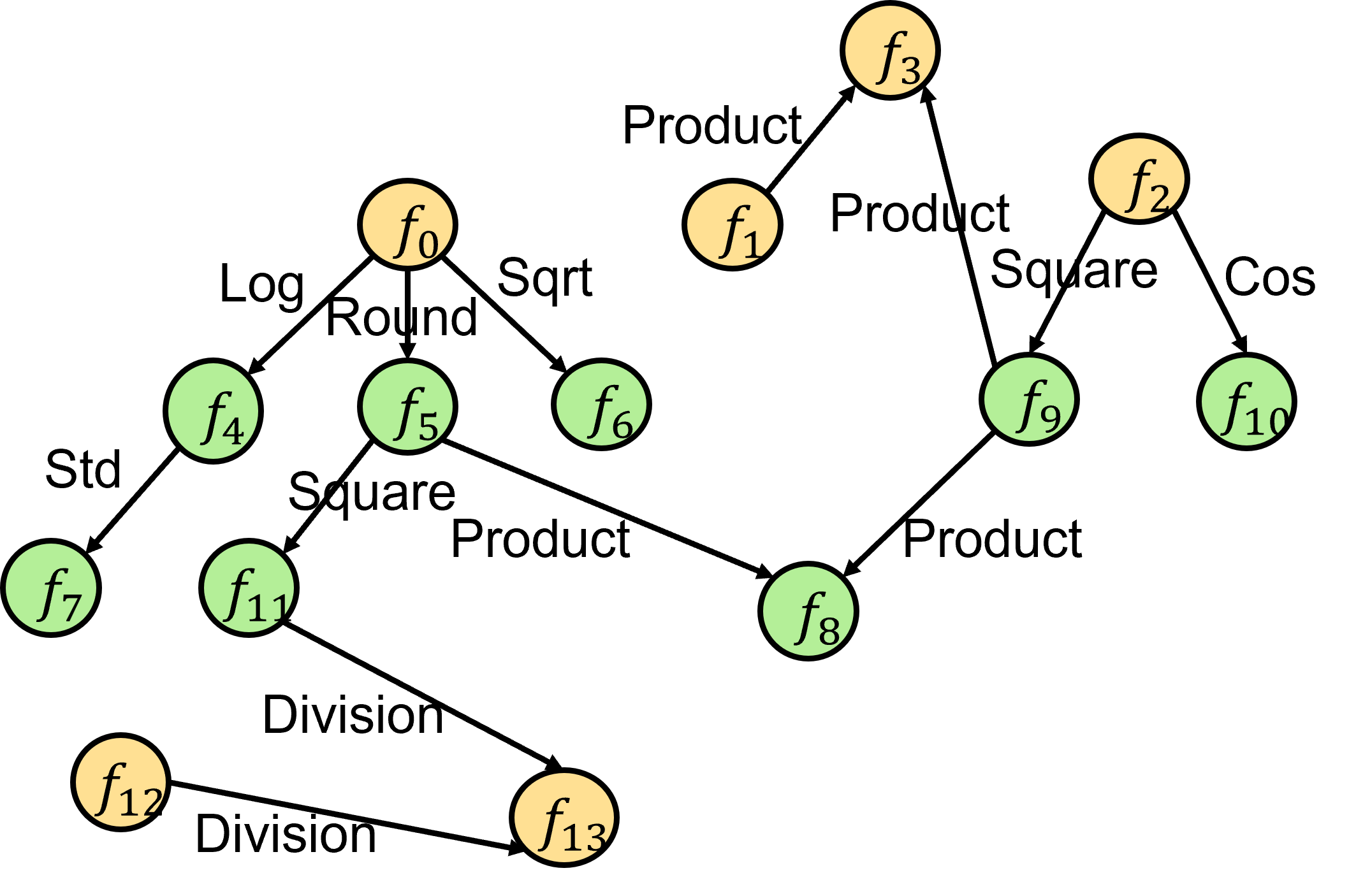}
        \caption{Decomposition Graph}
        \label{fig:dg}
    \end{subfigure}

    \caption{An overview of the KGs used by \textit{SMART}.}
    \label{fig:KGs} 
\end{figure}

\subsubsection{Knowledge representation} 
The KG, displayed in Figure \ref{fig:kg}, contains two types of knowledge: (i) Domain-agnostic knowledge which includes knowledge about \textit{units of measurement} and \textit{quantities} from various domains (e.g. Physics, Geometry, the International System of Units), and semantics about arithmetic and aggregation transformations; (ii) Domain-specific knowledge that covers concepts from various business domains, such as Healthcare, Banking, Retail, E-commerce, Finance, and others. This part of the KG can be easily expanded over time to cover additional application areas using KG integration tools.  


\subsubsection{Feature generation}
As stated before, \textit{SMART} exploits the KG to generate new domain-specific features through symbolic reasoning. More specifically, we use HermiT \cite{shearer2008hermit}, a reasoner for the DL syntax, to infer new knowledge based on the logical relationships of the KG. We display in Table \ref{tab:Tbox} a toy example of the \textit{Terminology box} (TBox) used in our experiments (due to privacy matters). For instance, if our dataset has a feature \textit{Date}, we can use the $7^{th}$ rule in Table \ref{tab:Tbox} to infer three new features: \textit{Day}, \textit{Month}, and \textit{Year}. We also can infer \textit{Population Total} from \textit{Location} or from \textit{City} features. Similarly, In the field of physics, the reasoner can infer \textit{Energy} for an entity that has a \textit{Mass} and a\textit{Velocity}. In the domain of banking and insurance, we can infer \textit{LoanApproval} from \textit{credit score} and \textit{income} of the loan applicant. In addition, we used SWRL (Semantic Web Rule Language) to define specific rules to determine whether a feature is interpretable. We show an example below. For instance, the first rule states that adding two features with different units would result in a non-interpretable feature and that periodic inventory totals are not summable. Similar explanations could be given to the other rules. 

\begin{align*}
(1)\text{ ~ } & Feature(?x) \land hasUnit(?x, ?u) \land Feature(?y) \land  \\
& hasUnit(?y, ?v) \land Feature(?z) \land Addition(?f) \land   \\
& hasInput(?f, ?x) \land hasInput(?f, ?y) \land Different(?u, ?v)  \\
& \land hasOutput(?f, ?z) \rightarrow nonInterpretable(?z)\\
(2)\text{ ~ }  & aggregationSum(?f) \land Stock(?x) \land hasInput(?f,?x) \\
&\land Feature(?z) \land hasOutput(?f,?z)\\
& \rightarrow nonInterpretable(?z)\\
\end{align*}

Furthermore, we use the relationships between entities in the KG to extract new features, by looking at the relations where a feature participates as a subject or an object, i.e. incoming and outgoing edges in the KG. For example, using the feature \textit{City} and relation \textit{locatedIn}, we can generate features for the entities that are located in the city, such as \textit{Universities}, \textit{Companies} or even important \textit{Events}. These features may have a significant impact on the ML model.   

\begin{table}[ht!]
\centering
\caption{Sample of the TBox}
\label{tab:Tbox}
\begin{tabular}{| l l l |} 
 \hline
 Function & $\sqsubseteq$ & $\geq 1$ $hasInput.(\forall hasUnit.Units)$ $\sqcap$ \\
 & & $\geq 1$ $hasOutput.(\forall hasUnit.Units)$ \\
  
 Arithmetic & $\sqsubseteq$ & Function $\sqcap$ $\leq 2$ $hasInput$ $\sqcap$ $\leq 1$ $hasOutput$ \\ 

 Addition & $\sqsubseteq$ & Arithmetic $\sqcap$ $\geq 2$ $hasInput.(\forall hasUnit.Unit$) $\sqcap$  \\
 & & $\leq 1$ $hasOutput.(\forall hasUnit.Unit$) \\
 
 Cos & $\sqsubseteq$ & $\leq 1$ $hasInput.(\forall hasUnit.angleU$) $\sqcap$ \\
 & & $\leq 1$ $hasOutput.Double$ \\

 Sin & $\sqsubseteq$ & $\leq 1$ $hasInput.(\forall hasUnit.angleU$) $\sqcap$ \\
 & & $\leq 1$ $hasOutput.Double$ \\

 ... & & \\
 
 $\bot$ & $\sqsupset$ & Feature $\sqcap$ Function \\
 
 Date & $\sqsubseteq$ & $\exists hasDay.Day$ $\sqcap$ $\exists hasMonth.Month$$\sqcap$ \\
 & & $\exists hasYear.Year$\\

Location & $\sqsubseteq$ &  $\exists hasCountry.Country$ $\sqcap$ $\exists hasCity.City$\\

Energy & $\sqsubseteq$ & $\exists hasMass.Mass$ $\sqcap$ $\exists hasVelocity.Velocity$\\

LoanApproval & $\sqsubseteq$ & Customer $\sqcap$ $\exists hasIncome.Income$ $\sqcap$ \\
& & $\exists hasGoodCreditScore.Score$\\

PremiumInsured & $\sqsubseteq$ & Customer $\sqcap$ $\exists hasGoodHealth.Health$ $\sqcap$ \\
& & $\exists hasYoungAge.Age$ ...\\


\hline
\end{tabular}
\end{table}

%% file: Exploration.tex
\subsection{Exploration}
\label{sec:exploration}
The goal of this step is to learn an exploration strategy of the search space, to generate interpretable features that may not be explicitly present in the KG. In the beginning, the search space is given by the following equation:
\begin{equation}
    \label{eq:searchSpace}
    Dom^F = \bigcup_{i=1}^{p} \text{ }\Biggl\{ \Biggl\{ \bigcup_{1 \leq s_1 \leq ..\leq s_i \leq p} \{ (f_{s_1},..,f_{s_i})\} \Biggl\} \times 
 \text{ } t_i \Biggl \},
\end{equation}
where $p$ is the number of features and $t_i \subseteq \mathcal{T}$ is the set of $i$-ary transformations. The number of elements of this space is: $|Dom^F| = \sum_{i=1}^{p} (A_i^p \times |t_i|)$, where $A_i^p$ is the $i$-permutation of $p$ features. It grows exponentially, hence an exhaustive search is not feasible. Therefore, \textit{SMART} employs an RL agent, as illustrated in Figure \ref{fig:Archi}, to generate new features without exploring all the search space. The core of this step is a policy network, instantiated by a well-designed deep neural network model, that takes the previously augmented dataset, or a bootstrapped counterpart to cope with the scaled dataset, as a state, and produces a FE pipeline through an inference pass.

\subsubsection{Feature generation with DQN.}
As discussed above, FE process conforms well to the trial-and-error process. Therefore, MDP can be used to automate the simulation of the FE control problem. Next, we define the underlying concepts of MDP in our context.

\textbf{State.} One key innovation of \textit{SMART} is our knowledge-driven MDP setup where it maps the tabular data, prior- and post- transformation, into the state using a semantic vectorization. Basically, each feature $x$ in the dataset is represented with a feature vector, $\Phi(x)$. Each dimension of this vector corresponds to a concept within the KG. The dimensions relating to $x$ are set to $1$, while the other dimensions are set to $0$. Since the number of features is different at each iteration, a final feature vector, $\Phi(X)$, is created by summing the feature vectors.

\textbf{Action.} The set of actions, $A$, corresponds to the set of transformations $\mathcal{T}$. We distinguish different types of transformations: (i) Unary functions: such as \textit{logarithm} and \textit{square root} for numerical features and \textit{One-Hot-Encoding} for categorical features; (ii) Binary functions: like arithmetic operators ($+, -, \times, \div$) and logical operators ($\land, \lor,...$); (iii) Aggregation functions (e.g. GroupByMean, GroupBySum) (iv) Date operators: such as \textit{Is\_Weekend}, \textit{Is\_Rush\_Hour}, to extract important information from dates. When an action $a_t \in A$ is applied on state $s_t$, a new state $s_{t+1} = \{s_t, a_t(s_t)\}$ is generated by concatenating the data of the previous state $s_t$ with $a_t(s_t)$.

\textbf{Reward.} Our reward calculation process consists of a linear combination of two metrics. The first one is and evaluation metric $\mathcal{P}$, (e.g. RMSE, AUC), related to a pre-selected ML model, $L$ (e.g. XGBoost). It consists of the average performance gained from the previous step on a k-fold cross validation, i.e $\mathcal{P}(L(\hat{X}_i,Y)) - \mathcal{P}(L(\hat{X}_{i-1},Y))$. The second metric, introduced in the following Section \ref{inter_metric}, is used to asses features interpretability.

\textbf{Selecting an action.} To select an action, we used \textit{decaying $\epsilon$-greedy} \cite{wunder2010classes}. It consists of taking the action with the maximum reward with a probability of $1-\epsilon$, or a random action with a probability of $\epsilon$, while decreasing $\epsilon$ overtime. Thus, in the beginning, we encourage the agent to explore its environment. Once it learns, we decrease $\epsilon$, allowing the agent to exploit its knowledge.

\textbf{Policy.} DQN is modeled by a policy network $\pi : S \rightarrow Q(A)$, where $Q$ is the expected cumulative reward estimated by the agent. The goal is to obtain the best sequence of $m$ transformations. We used two instances of a fully-connected multi-layer neural network to approximate the Q-function: (i) The main neural network, $Q(s,a;\theta_i)$, takes the current state and outputs an expected cumulative reward, $Q-$value, for each possible action; (ii) The target network, $\hat{Q}(s',a';\hat{\theta}_i)$, with delayed updates, helps mitigating issues related to moving target distributions. A Q-learning update at iteration $i$ is defined as the loss function: 
\begin{equation}
    \label{loss}
    L_i(\theta_i) = \mathbb{E}_{(s,a,r,s') \sim U(D)}\left[\left(r +\gamma \max_{a'} \hat{Q}(s',a';\hat{\theta_i}) - Q(s,a;\theta_i)\right)^2 \right]
\end{equation}
We stored past experiences in a replay buffer and sampled them during training. This helps break the temporal correlation between consecutive samples, providing a more stable learning process.

Basically, at each step of the training, and iteratively until convergence, the agent receives the state representing the input dataset and feed it to a neural network. This latter estimates an intermediate reward score for each possible function that can be applied and selects the one that maximizes the long term reward. The selected function is then used to generate new features.

\subsubsection{Feature interpretability metric.} 
\label{inter_metric}
Despite the acknowledgement of the importance of feature interpretability, there is no quantitative metric to evaluate it \cite{zytek2022need}. Therefore, in this work, we propose a new metric to assess the interpretability of a feature by calculating the distance between this feature and the concepts of a graph called Decomposition Graph (\textit{DecomG}). Each node in \textit{DecomG} represents a feature. It could be either a known concept inferred from the KG during the exploitation, or a new feature generated by the RL agent. The edges correspond to the transformations. An edge, $t_k$ from a feature $x_i \in \hat{X}_\mathcal{T}$ to a feature $x_j \in \hat{X}_\mathcal{T}$ indicates that $x_j$ is derived from $x_i$, i.e. $x_j = t_k(x_i)$. 

To compute the interpretability of a generated feature $\Tilde{x}$, we first identify all paths within \textit{DecomG} that directly connect $\Tilde{x}$ to known concepts (features generated during exploitation). Each path $p_k$ is assessed based on the interpretability of the features linked to $\Tilde{x}$ in this path, $\{x_i\}$, and the complexity of the transformations applied, $T=\{t_j\}$. The interpretability score for a path, $\mathcal{I}_{KG}^{p_k}(\Tilde{x})$, is calculated recursively as the product of the interpretability scores of its transformation and the least interpretable feature along the path, as shown in Equation \ref{eq:inter1}. The overall interpretability of $\Tilde{x}$, denoted as $\mathcal{I}_{KG}(\Tilde{x})$, is then determined by the maximum interpretability score among all computed paths (Equation \ref{eq:inter2}).

\begin{equation}
    \label{eq:inter1}
        \mathit{\mathcal{I}_{KG}^{p_k}(\Tilde{x}) = Inter(t_j) \times \argmin\{\mathcal{I}_{KG}(x_i)\}} \\ 
\end{equation}
\begin{equation}
        \label{eq:inter2}
            \mathit{\mathcal{I}_{KG}(\Tilde{x}) = \argmax_{p_k}\{I_{KG}^{p_k}(\Tilde{x})\}} \\ 
\end{equation}

%% file: Exp.tex
\section{Analysis and Interpretation}
In our experiments we aim to address the following key research questions: 

\begin{itemize}

    \item \textbf{Q1:} How effective is our approach compared to the state-of-the-art?
    \item  \textbf{Q2:} How good is the interpretability of the features generated by \textit{SMART}?

\end{itemize}
    
\subsection{Experimental Settings}

\subsubsection*{\underline{\textbf{Datasets}}}
We used several widely-used real-world datasets that covered various data characteristics, e.g., domains, size and number of features. We describe some of them in the following.

\begin{enumerate}
     \item \textit{Wine Quality White/Red} \cite{CorCer09} are used to model wine quality based on physico-chemical tests. They can be seen as a classification of regression problem.

     \item \textit{Bikeshare DC} \cite{bike} contains the hourly and daily count of rental bikes between $2011$ and $2012$ in Capital bikeshare system in Washington DC with the corresponding weather and seasonal information.
    
    \item \textit{Home Credit Default Risk} contains information on loan repayments.

    \item \textit{Amazon Employee} predicts if an amazon employee is allowed/denied access to some resources based on his role and the characteristics of the required resource. This dataset is part of \textit{Amazon.com - Employee Access Challenge}.

    

    \item \textit{Higgs Boson} \cite{baldi2014searching} is a classification dataset to distinguish between a signal process which produces Higgs bosons and a background process which does not.
    
    \item \textit{Brazilian E-Commerce} \cite{brazil} is a dataset with a multi classiffication tasks to predict the score of an order given by the customer

    \item \textit{Medical appointment} predicts if a patient who made a doctor appointment would show up or not. 

    \item \textit{Fraud Detection} is used to predict the probability that an online transaction is fraudulent.

    \item \textit{Retail Spending} is used to make a churn prediction model.

    \item \textit{NYC Taxi Ride Data} \cite{chepurko2020arda} is used to predict the total ride duration of taxi trips in New York City.
\end{enumerate}

\subsubsection*{\underline{\textbf{Baselines}}} We compared \textit{SMART} with several methods. 

\begin{itemize}
    \item \textbf{Base} is the original dataset without FE. 
    
    \item  \textbf{Random} consists of randomly applying transformations to raw features to generate new ones. 
    
    \item \textbf{Deep Feature Synthesis (DFS)} applies all transformations on all features then performs feature selection \cite{kanter2015deep}. 
    
    \item \textbf{Reinforcement Learning based model (RLM)} builds a transformation graph and employs Q-learning techniques \cite{khurana2018feature}. 
    
    
    \item \textbf{mCAFE} \cite{huang2022automatic} is an RL method based on monte carlo tree search and an LSTM neural network for the expansion policy. 
    
    \item \textbf{NFS} \cite{chen2019neural} is a neural architecture search based approach that utilizes several RNN-based controllers to generate transformation sequences for each raw feature. 
    
    
    \item \textbf{DIFER} \cite{zhu2022difer} is a feature evolution framework based on the encoder-predictor-decoder.
\end{itemize}

\subsubsection*{\underline{\textbf{Evaluation Metrics}}}
For effectiveness, we use different evaluation metrics for different tasks. We use the metric \textit{1 - rae} (relative absolute error) \cite{shcherbakov2013survey} for regression (Reg.) problems:
\begin{equation}
    \label{eq:metrics}
    1-rae = 1 - \dfrac{\sum{|\hat{y}-y|}}{\sum{|\bar{y}-y|}}
\end{equation}
where $y$ is the target, $\hat{y}$ is the model prediction and $\bar{y}$ is the mean of $y$. For classification (Class.) tasks, following \cite{khurana2018feature}, we use \textit{F1-score} (the harmonic mean of precision and recall).

\subsection{Effectiveness of \textit{SMART}.}

\begin{table*}[htbp]
    \centering
    \caption{Comparing the effectiveness of \textit{SMART} with the baselines}
    \label{tab:all}
    \resizebox{1\linewidth}{!}{
    \begin{tabular}{c c c c c c c c c c|c c c c c c c c c c}
        \hline
        Dataset & ALG & Base & Random & DFS & AutoFeat & NFS & DIFER & mCAFE & SMART &Dataset & ALG & Base & Random & DFS & AutoFeat & NFS & DIFER & mCAFE & SMART\\
        \hline
        
        \multirow{5}{*}{\textbf{Diabetes}} & RF & 0.740  & 0.670 & 0.737 & 0.767 & 0.786 & 0.798 & 0.813 & \textbf{0.853} & \multirow{5}{*}{\textbf{Medical}} & RF & 0.491 &  0.484 &  0.499 & 0.790 & 0.650 & 0.779 & 0.786 & \textbf{0.893} \\
        & DT & 0.732 & 0.598 & 0.732 & 0.741 & 0.770 & 0.776 & 0.798 & \textbf{0.849} & & DT & 0.478 & 0.480 & 0.487 & 0.780 & 0.634  & 0.729 & 0.753 & \textbf{0.893}\\
        & LR & 0.753 & 0.613 & 0.748 & 0.780 & 0.792 & 0.816 & 0.819 & \textbf{0.857} & & LR & 0.502 & 0.498 & 0.517 & 0.803 & 0.702 & 0.829 & 0.753 & \textbf{0.897} \\
        & SVM & 0.742 & 0.647 & 0.719 & 0.756 & 0.762 & 0.770 & 0.788 & \textbf{0.850} & \textbf{appointment} & SVM & 0.482 & 0.476 & 0.491 & 0.782 & 0.641 & 0.739 & 0.764 & \textbf{0.891} \\
        & XGB & 0.755 & 0.732 & 0.750 & 0.788 & 0.792 & 0. 813 & 0.819 & \textbf{0.863} & & XGB & 0.503 &0.503  & 0.515 & 0.801 & 0.713 & 0.830 & 0.838 & \textbf{0.892}\\
        \hline 

        \multirow{5}{*}{\textbf{German Credit}} & RF & 0.661 & 0.655 & 0.680 & 0.760 & \textbf{0.781} & 0.777 & 0.770 &\textbf{0.781}&  \multirow{5}{*}{\textbf{Home Credit}} & RF & 0.797 & 0.766 & 0.802 & 0.806  & 0.799 & 0.810 & 0.801 & \textbf{0.877}\\
        & DT & 0.651 & 0.650 & 0.672 & 0.720 & 0.761 & 0.752 & 0.765 & \textbf{0.769 }& & DT & 0.782 & 0.761 & 0.789 &0.7.92 & 0.792 & 0.802 & 0.789 & \textbf{0.877} \\
        & LR & 0.670 & 0.657 & 0.692& 0.767 &0.788 & 0.773 & 0.782 & \textbf{0.790} & & LR & 0.801 & 0.783 & 0.804 & 0.805 &  0.807 & 0.811 & 0.805 & \textbf{0.882}\\
       & SVM & 0.655 & 0.661 & 0.679 & 0.734 & 0.765 & 0.762 & 0.772 & \textbf{0.780}  & \textbf{Default Risk} & SVM & 0.789 & 0.769& 0.793& 0.790& 0.801 & 0.806 & 0.803 & \textbf{0.878} \\
        & XGB & 0.690 & 0.666 & 0.695 & 0.770 & 0.790 & 0.792 & 0.790 & \textbf{0.795} & & XGB &0.802 & 0.780  & 0.804 & 0.811 & 0.819 & 0.814 &0.809 & \textbf{0.885}\\
        \hline
        
        \multirow{5}{*}{\textbf{Wine Quality}} & RF & 0.689 & 0.678 & 0.654 & 0.502 & 0.707 & 0.515 & 0.502 &\textbf{0.722}&  \multirow{5}{*}{\textbf{Openml\_618}} & RF & 0.428 & 0.428 & 0.411 & 0.632 & 0.640 & 0.660 & 0.738 & \textbf{0.743}\\
        & DT & 0.492 & 0.473 & 0.475 &0.489 &   0.513 & 0.513 & 0.492 & \textbf{0.713 }& & DT & 0.425 & 0.419 & 0.408 & 0.630 &    0.629 & 0.645 & 0.725& \textbf{0.737} \\
        & LR & 0.501  & 0.489 & 0.490 &0.505 &    0.519& 0.517 & 0.509 & \textbf{0.727} & & LR &0.432 & 0.429&0.412 &0.645 &   0.642 & 0.666& 0.740& \textbf{0.749}\\
        \textbf{White}& SVM & 0.489 & 0.480 & 0.482 & 0.493 &   0.521 & 0.515& 0.499& \textbf{0.719} & & SVM & 0.423& 0.421& 0.405 & 0.635 &   0.632 & 0.643 & 0.720 & \textbf{0.738} \\
        & XGB & 0.503& 0.490& 0.493& 0.508 &   0.525&0.524 &0.511 & \textbf{0.735} & & XGB & 0.430 & 0.429 & 0.413 & 0.642 &   0.645 & 0.663 & 0.735 & \textbf{0.743}\\
        \hline

        \multirow{5}{*}{\textbf{Amazon}} & RF & 0.712 & 0.740 & 0.744 & 0.739 & \textbf{0.945} & 0.909 & 0.897 & 0.912 & \multirow{5}{*}{\textbf{Bikeshare DC}} & RF & 0.393 & 0.381 & 0.693 & 0.849 & 0.974 & 0.981 & 0.906 & \textbf{0.988}\\
        & DT & 0.701 & 0.736 & 0.738 & 0.732 &   \textbf{0.932} & 0.903 & 0.878 & 0.910 & & DT & 0.378 & 0.372 & 0.682 & 0.840 &   0.965 & 0.959 & 0.896 & \textbf{0.972} \\
        & LR & 0.715 & 0.744 & 0.750 &0.742 &   \textbf{0.935}& 0.913 & 0.905 & 0.925 & & LR & 0.401 & 0.385 & 0.702 & 0.853 &   0.975 & 0.981 & 0.921 & \textbf{0.988}\\
        \textbf{Employee}& SVM & 0.710 & 0.742 &0.740 & 0.735&   \textbf{0.936} & 0.910& 0.906 & 0.932& & SVM & 0.380& 0.381& 0.695& 0.845&  0.978 & 0.978&0.922 &\textbf{0.980} \\
        & XGB &0.719 &0.748 &0.755 &0.745 &   \textbf{0.945} & 0.915& 0.908& 0.930 & & XGB & 0.405& 0.389& 0.712& 0.855&   0.980& 0.982& 0.932& \textbf{0.991}\\
        \hline

        \multirow{5}{*}{\textbf{Higgs Boson}} & RF & 0.718 & 0.699 & 0.682 & 0.468 & 0.731 & 0.738 & 0.739 & \textbf{0.743} & \multirow{5}{*}{\textbf{NYC Taxi}} & RF &  0.425 & 0.434 & 0.505  & - & 0.551 & 0.501 & 0.446 & \textbf{0.610}\\
        & DT & 0.709& 0.684& 0.678& 0.460&  0.725 &0.719 & 0.728&\textbf{0.730} & & DT & 0.419 & 0.429 & 0.489 & -& 0.539& 0.487& 0.426& \textbf{0.608}\\
        & LR & 0.722& 0.704& 0.700& 0.470&  0.734 &0.742 &\textbf{0.745} & 0.739 & &  LR & 0.430& 0.449 & 0.513 & 0.458 & 0.559 & 0.515 & 0.452 &\textbf{0.614} \\
        \textbf{Ride}& SVM &0.711 &0.705 &0.689 &0.465 &  0.726 & 0.724& \textbf{0.732}& \textbf{0.732}& & SVM & 0.423 & 0.432 & 0.499 & - & 0.545 & 0.492 & 0.459& \textbf{0.615}\\
        & XGB &0.725 & 0.705& 0.706& 0.489&   0.735& 0.740& 0.735&\textbf{0.745} & & XGB & 0.432 & 0.459 & 0.515 & - & 0.562 & 0.523 & 0.486 & \textbf{0.623} \\
        \hline
    \end{tabular}
}
\end{table*}

We report in Table \ref{tab:all} the comparison of 10 datasets  (out of 14 due to lack of space). Despite having the interpretability constraint, our approach outperforms the baselines in most cases. Compared to raw data, the features generated by \textit{SMART} can improve the performance by an average of $20.94\%$. Compared to \textit{DIFER}, \textit{NFS} and \textit{mCAFE}, \textit{SMART} achieves an average improvement of $11.55\%$, $4.86\%$ and $7.24\%$ respectively. Notably, \textit{SMART} can handle different number of instances and features. 

The performance of our model can be attributed to several key factors. Firstly, it leverages domain knowledge to generate features with rich semantics that would be difficult to discover directly from the data. By incorporating this domain understanding, \textit{SMART} generates features that better explain the target variable, which aligns with the core concept of FE: enriching the dataset with meaningful features. Moreover, \textit{SMART}'s ability to generate high-order features plays an important role in its performance. Moreover, \textit{SMART}'s ability to generate high-order features plays an important role in its performance. In fact, the composition of transformations is crucial for discovering complex and interesting relationships between features. Additionally, unlike most baselines that rely on basic arithmetic functions, \textit{SMART} implements a wider range of transformations, including aggregations and logical operators. This variety allows the model to generate more promising features, ultimately leading to better performance.

\subsection{Interpretability Evaluation of \textit{SMART}.}

In our work, we focus on feature interpretability for domain experts, rather than the model interpretability. Our experience showed that even simple and interpretable models, such as regression, can become challenging to understand if they rely on non-interpretable features. This emphasizes the importance of interpretability at the feature level to enhance the overall interpretability of ML models. 

\begin{figure}
    \centering
    \captionsetup[subfigure]{labelformat=empty}
    
    \begin{subfigure}{0.49\linewidth}
        \centering
        \includegraphics[width=\linewidth, height=3cm]{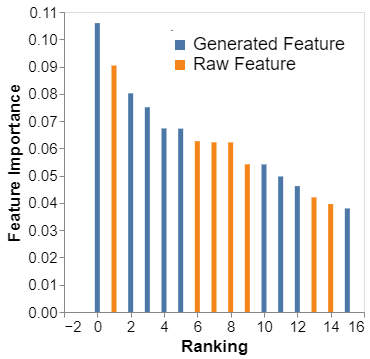}
        \caption{Diabetes}
        \label{fig:diabete}
    \end{subfigure}
    \hfill
    \begin{subfigure}{0.49\linewidth}
        \centering
        \includegraphics[width=\linewidth, height=3cm]{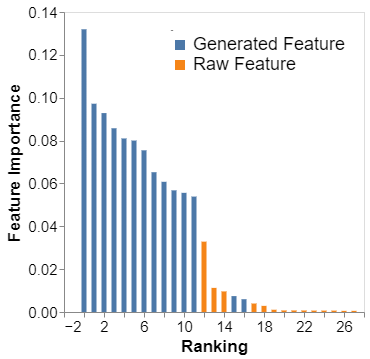}
        \caption{NYC Taxi}
        \label{fig:nyc}
    \end{subfigure}
    \hfill
    \begin{subfigure}{0.49\linewidth}
        \centering
        \includegraphics[width=\linewidth, height=3cm]{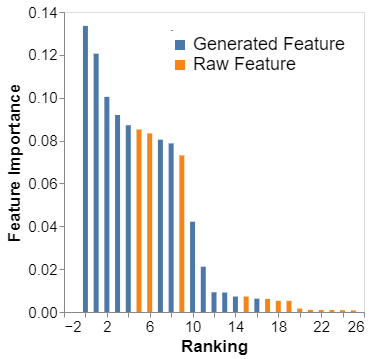}
        \caption{Medical Appt}
        \label{fig:med}
    \end{subfigure}
    \hfill
    \begin{subfigure}{0.49\linewidth}
        \centering
        \includegraphics[width=\linewidth, height=3cm]{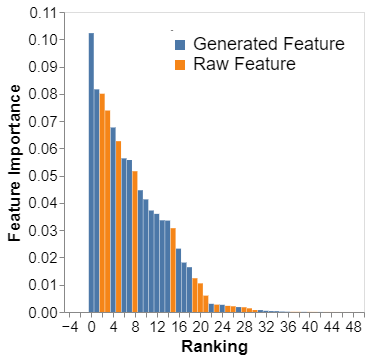}
        \caption{Credit Risk}
        \label{fig:home}
    \end{subfigure}

    \caption{Feature Importance.}
    \label{fig:featuresImpranking} 
\end{figure}

\textbf{Interpretability based on training data.} From the field of XAI, data-based interpretability attributes ML predictions to specific parts of the training data. In this section we use the well-known approach SHAP (SHapley Additive exPlanations) \cite{lundberg2017unified} to quantify the contribution of input features to the model prediction. Then, we compare feature importance of raw features with features generated by our model. To conduct the experiment, we create a new dataset by combining the $n$ raw features with the top-ranked $n$ features generated by \textit{SMART}. We then use SHAP to score feature importance. Figure \ref{fig:featuresImpranking} showcases the results for four different datasets. Notably, the features generated by \textit{SMART} (shown in blue) are more important compared to raw features (shown in orange) across all datasets, thereby validating the effectiveness of \textit{SMART} in generating interpretable features.

\begin{figure*}[ht!]
    \centering
    \captionsetup[subfigure]{labelformat=empty}
    \begin{subfigure}{0.3\linewidth}
        \centering
        \includegraphics[width=\linewidth]{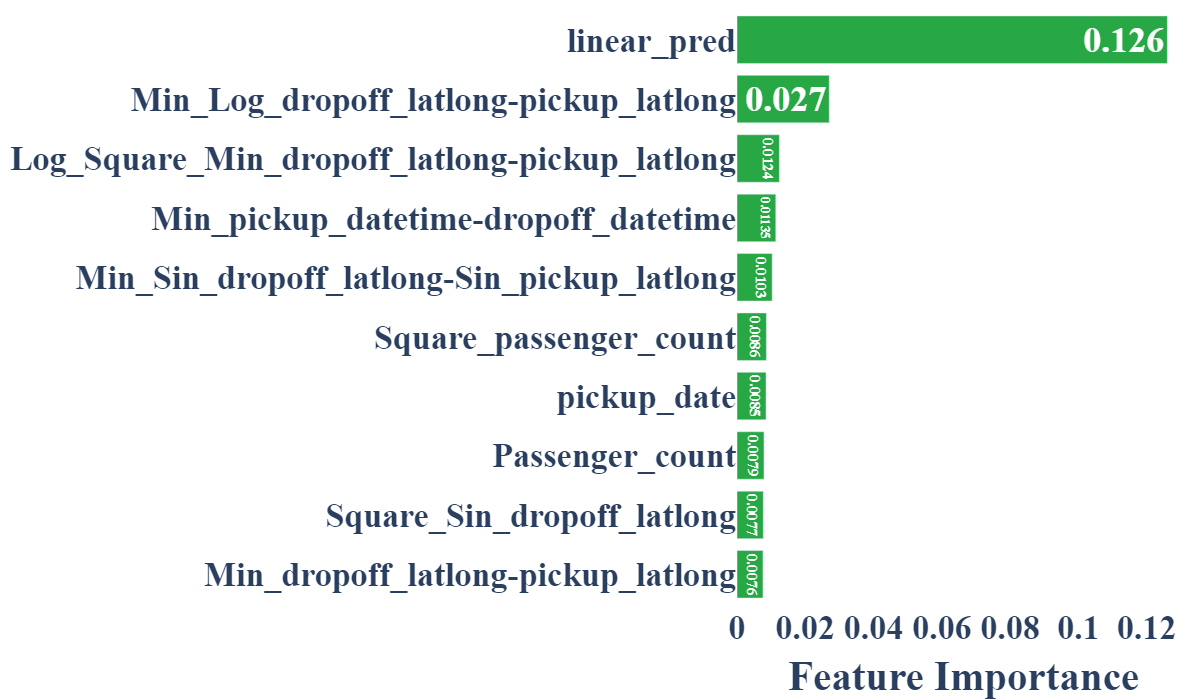}
    \end{subfigure}
    \hfill
    \begin{subfigure}{0.3\linewidth}
        \centering
        \includegraphics[width=\linewidth]{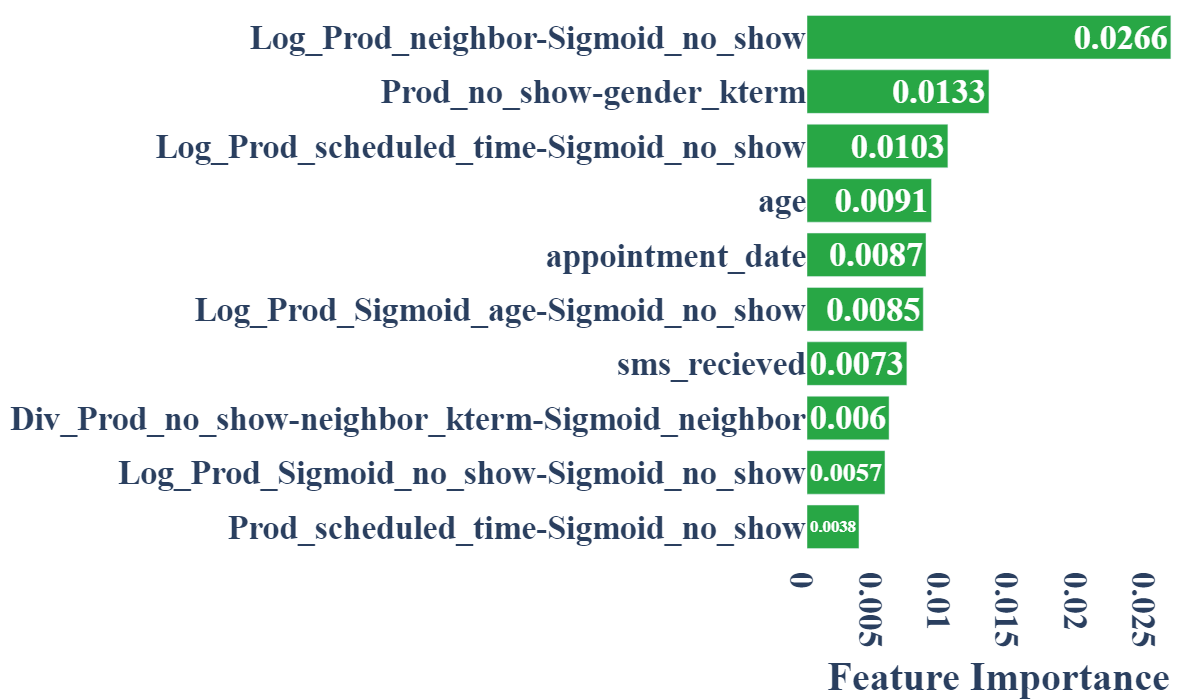}
    \end{subfigure}
    \hfill
    \begin{subfigure}{0.3\linewidth}
        \centering
        \includegraphics[width=\linewidth]{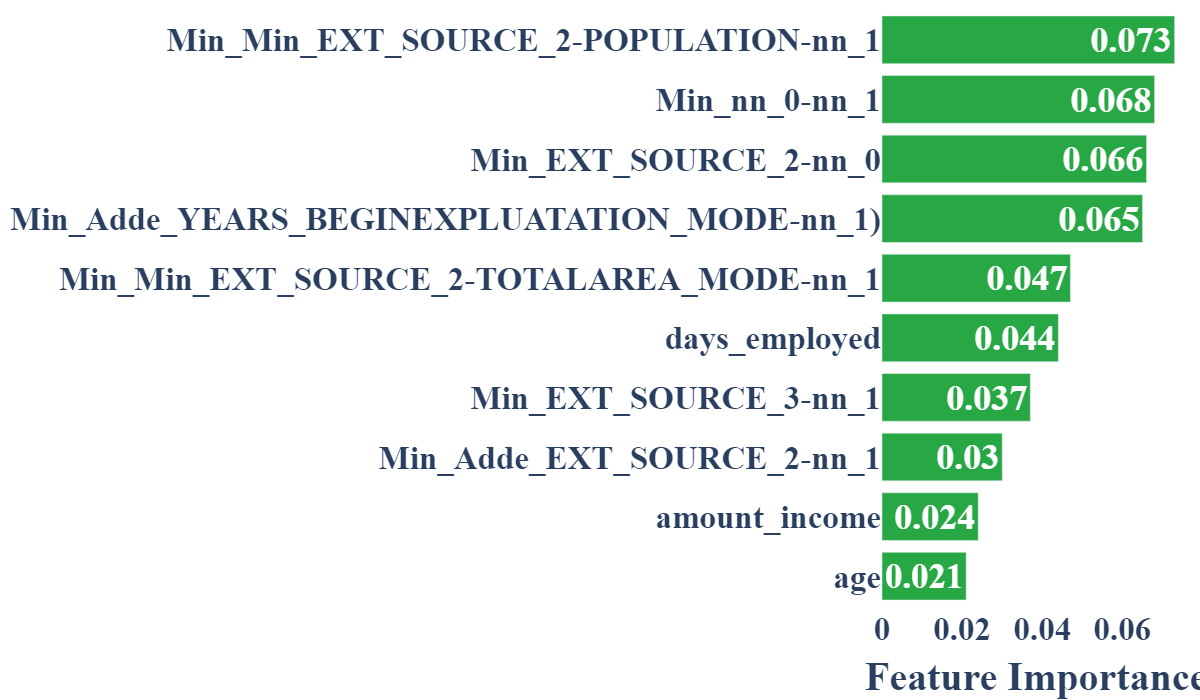}
    \end{subfigure}
    \\
    \begin{subfigure}{0.3\linewidth}
        \centering
        \includegraphics[width=\linewidth]{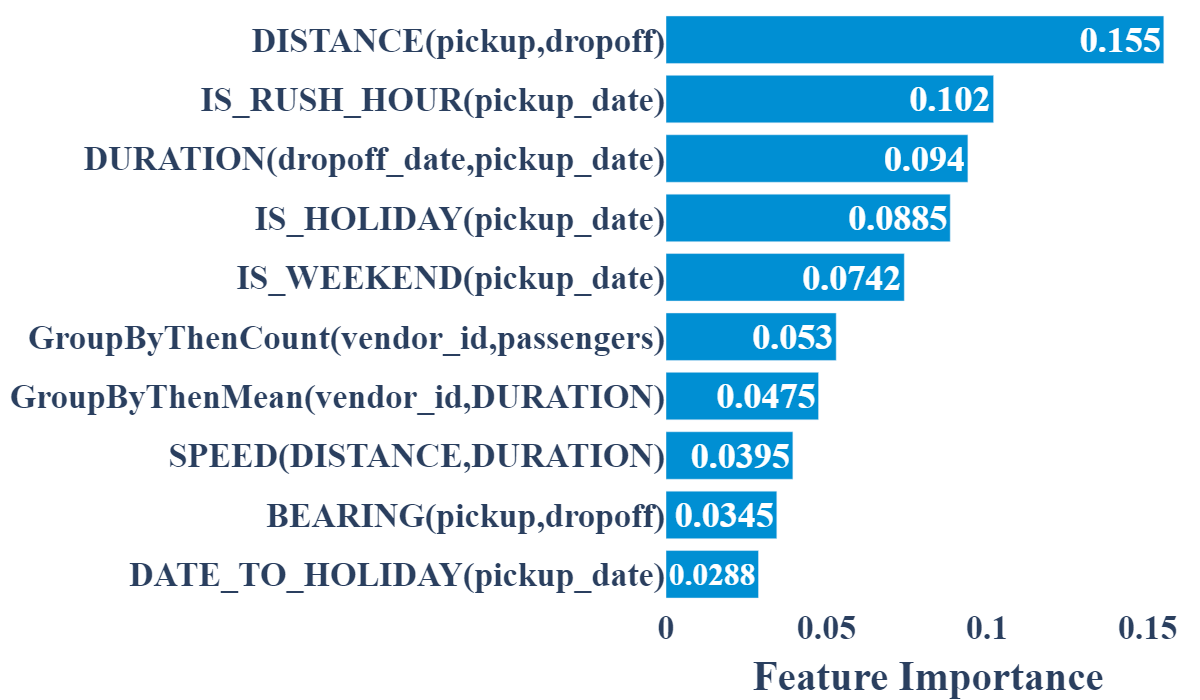}
        \caption{NYC Taxi Ride Duration}
        \label{fig:imp_nyc}
    \end{subfigure}
    \hfill
    \begin{subfigure}{0.3\linewidth}
        \centering
        \includegraphics[width=\linewidth]{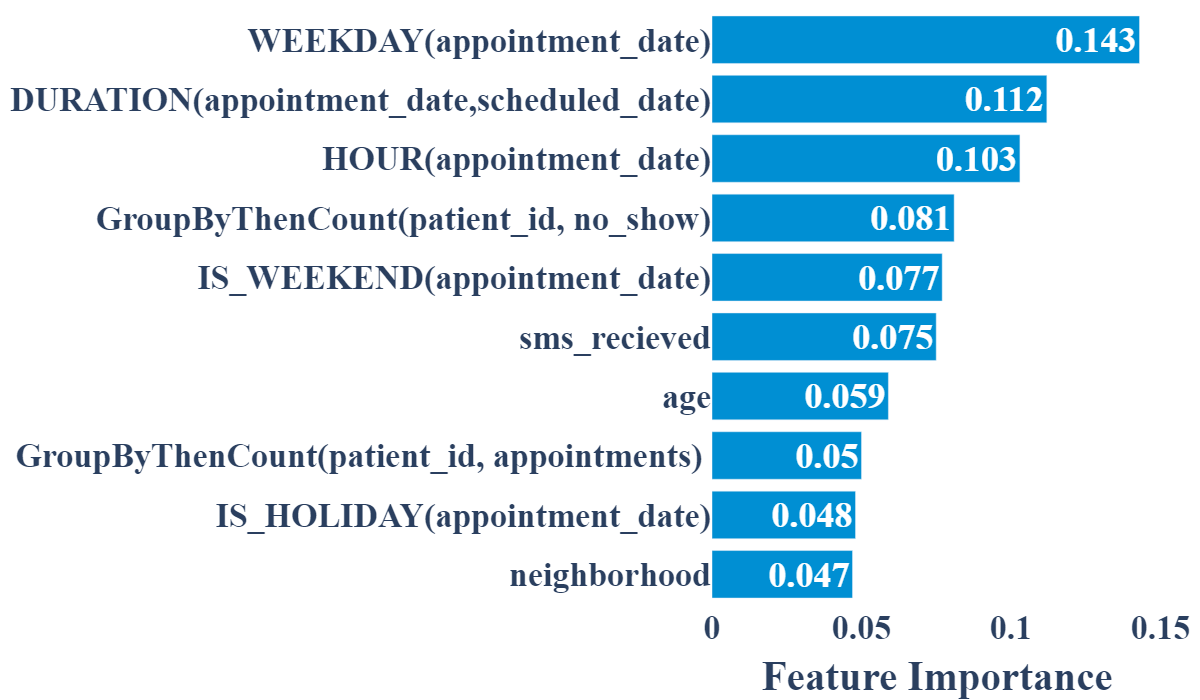}
        \caption{Medical Appointment}
        \label{fig:imp_med}
    \end{subfigure}
    \hfill
    \begin{subfigure}{0.3\linewidth}
        \centering
        \includegraphics[width=\linewidth]{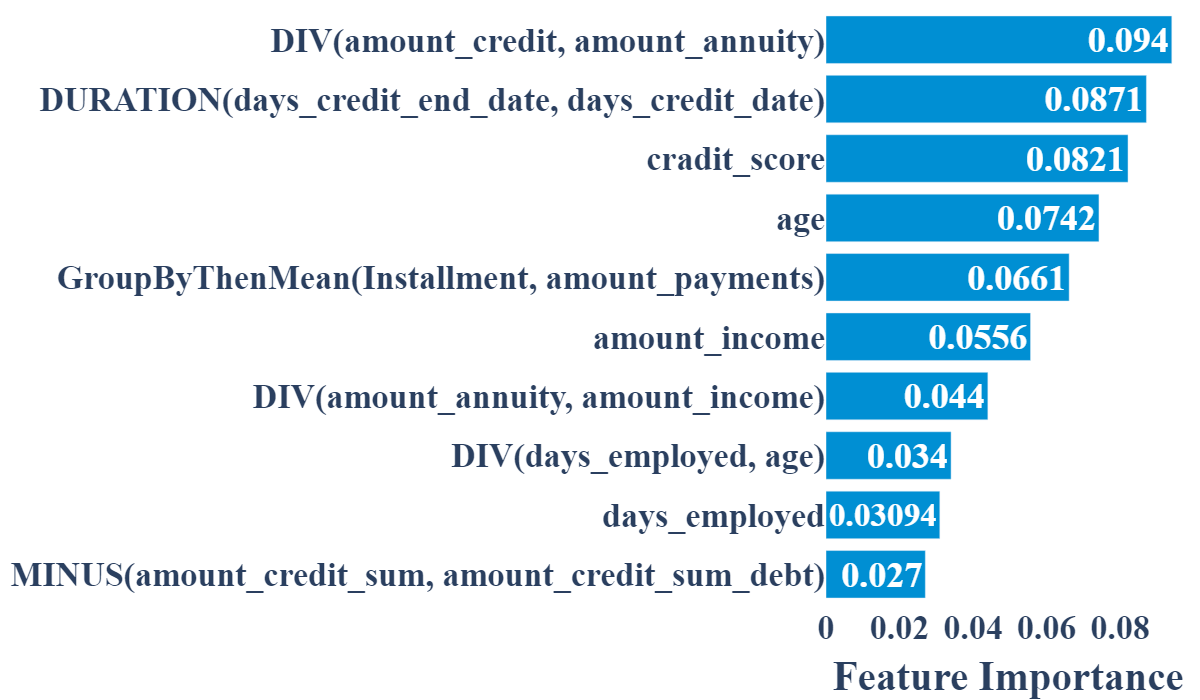}
        \caption{Credit Default Risk}
        \label{fig:imp_home}
    \end{subfigure}
    \caption{Top-10 features of \textit{mCAFE} (green) and \textit{SMART} (blue).}
    \label{fig:featuresImpo} 
\end{figure*}

\textbf{Feature Importance.} For a more in-depth analysis, we compare in Figure \ref{fig:featuresImpo} the top $10$ features of our model (in blue) with the ones generated by \textit{mCAFE} (in green) using three different datasets from different domains. It can be seen that, compared to the baselines, the features generated by our model are easily readable, enabling users to understand their meaning, and more interpretable, aligning with the domain knowledge. For instance, for \textit{Medical appointment} dataset, \textit{SMART} generated several useful features, including the duration between the date of the appointment and the actual appointment date. A thorough analysis of this dataset shows that as this duration increases, the likelihood of patients missing their appointments also increases. In the \textit{NYC Taxi Trip} dataset (Figure \ref{fig:imp_nyc}), our model generated $9$ out of the top $10$ features. Among them, the model used the \textit{Haversine} transformation to calculate the distance between pickup and dropoff locations, and calculated the duration between pickup and dropoff time showing a direct correlation with trip duration. However, the features generated by the baselines remain challenging for domain experts to understand, and often they do not relate to the domain knowledge.